\documentclass{article}
\usepackage{ijcai22}

\usepackage{times}
\usepackage{soul}
\usepackage{url}
\usepackage[hidelinks]{hyperref}
\usepackage[utf8]{inputenc}
\usepackage[small]{caption}
\usepackage{graphicx}
\usepackage{amsmath}
\usepackage{amsthm}
\usepackage{booktabs}
\usepackage{algorithm}
\usepackage{algorithmic}
\usepackage{multirow}

\usepackage[dvipsnames]{xcolor}

\newcommand{\revision}[1]{\textcolor{black}{#1}}

\title{Self-supervised Semantic Segmentation Grounded in Visual Concepts}

\author{
Wenbin He$^1$
\and
William Surmeier$^1$\and
Arvind Kumar Shekar$^{2}$\and
Liang Gou$^1$\and
Liu Ren$^1$
\affiliations
$^1$Robert Bosch Research and Technology Center North America\\
$^2$Robert Bosch GmbH
\emails
wenbin.he2@us.bosch.com,
clayton.surmeier@gmail.com,
arvindkumar.shekar@de.bosch.com,
liang.gou@us.bosch.com,
liu.ren@us.bosch.com
}

\begin{document}

\maketitle

\begin{abstract}

Unsupervised semantic segmentation requires assigning a label to every pixel without any human annotations.  Despite recent advances in self-supervised representation learning for individual images, unsupervised semantic segmentation with pixel-level representations is still a challenging task and remains underexplored.  In this work, we propose a self-supervised pixel representation learning method for semantic segmentation by using visual concepts (i.e., groups of pixels with semantic meanings, such as parts, objects, and scenes) extracted from images.  To guide self-supervised learning, we leverage three types of relationships between pixels and concepts, including the relationships between pixels and local concepts, local and global concepts, as well as the co-occurrence of concepts.  We evaluate the learned pixel embeddings and visual concepts on three datasets, including PASCAL VOC 2012, COCO 2017, and DAVIS 2017.  Our results show that the proposed method gains consistent and substantial improvements over recent unsupervised semantic segmentation approaches, and also demonstrate that visual concepts can reveal insights into image datasets.

\end{abstract}

\section{Introduction}
\label{sec:introduction}

Semantic segmentation plays a crucial role in a broad range of applications, including autonomous driving, medical image analysis, etc.  It partitions an image into semantically meaningful regions and assigns each region with a semantic label such as people, bikes, and cars.  Recently, semantic segmentation models~\cite{Zhao2017,Chen2018a} with deep convolutional neural networks (CNNs) have shown promising results on popular benchmarks.  However, these approaches rely heavily on pixel-wise annotations, which cost significant amounts of time and money to acquire.  Thus computer vision community starts to pay more attention to unsupervised/self-supervised approaches.

With recent advances in self-supervised learning, visual representations can be learned from images without additional supervision.  However, many self-supervised representation learning frameworks (e.g., SimCLR~\cite{Chen2020}, MOCO~\cite{He2020}) focus on the visual representation of a whole image and require curated single-object images.  Only a few recent approaches learn visual representations at the pixel level for complex scenes, including SegSort~\cite{Hwang2019}, Hierarchical Grouping~\cite{Zhang2020a}, and MaskContrast~\cite{VanGansbeke2021}.  These pixel-level representations can be used for semantic segmentation and show promising results.  The key idea of these approaches is to use contrastive learning to separate positive pixel pairs from negative pixel pairs.  They use different priors to formulate positive and negative pairs, such as contour detection~\cite{Hwang2019}, the hierarchical grouping of image segments~\cite{Zhang2020a}, and saliency masks~\cite{VanGansbeke2021}.  However, these methods largely emphasize pixels' local affinity information at individual image level but ignore the global semantics in the whole dataset, and thus tend to separate the representation of objects from different images even when these objects belong to the same semantic class.

In this work, we propose a self-supervised pixel representation learning method by leveraging some important properties of visual concepts at both local and global levels.  Visual concepts can be informally defined as human-interpretable abstractions of image segments with semantic meanings (e.g., parts and objects). The idea of visual concept is inspired by recent work in the field of eXplainable AI~\cite{Bau2017,Kim2018}, and offers us a unified framework to derive self-supervision priors at both local and global levels.

Specifically, we use three types of relationships between pixels and visual concepts to guide self-supervised learning.  First, aim to leverage the relationship between pixels and local concepts (i.e., visually coherent regions in each image), we use contrastive learning to force the pixels in the same pseudo segments moving close in the embedding space, and push pixels from either different pseudo segments or different images moving far away.  Secondly, we group local concepts with similar semantic meanings into global concepts with a vector quantization (VQ) method ~\cite{vandenOord2017}.  A VQ dictionary learns discrete representations of global concepts by pushing each local concept from an image to its closest global concept vector.  Lastly, we utilize the co-occurrence of different global concepts because relevant global concepts tend to appear in the same image, such as human face and body, or rider and bike.  These relationships based on visual concepts together regularize the self-supervised learning process.

The learned pixel embeddings can then be used for semantic segmentation by $k$-means clustering~\cite{Hwang2019} or fine-tuning~\cite{Zhang2020a,VanGansbeke2021}.  Moreover, our method can be used to extract visual concepts from a dataset and offer a global semantic understanding about the dataset.

In summary, the contributions of this paper are threefold:
\begin{itemize}
  \item We propose a self-supervised pixel representation learning method for semantic segmentation, which learns visual concepts for image segments and uses the relationships between pixels and concepts to improve the pixel representation.  Moreover, our method only uses pseudo image segments generated by weak priors (e.g., a contour detector or superpixels) without additional information such as hierarchical groups~\cite{Zhang2020a} or saliency masks~\cite{VanGansbeke2021}.
  \item Our approach can produce a set of global visual concepts that are semantically meaningful to humans.  Because there are a finite number of discrete visual vocabulary, they are easier to explore and interpret comparing with the high dimensional embeddings.
  \item We demonstrate the accuracy and generalizability of the learned pixel embeddings on three datasets, including PASCAL VOC 2012, COCO  2017, and DAVIS 2017.
\end{itemize}

\section{Related Work}
\label{sec:related_work}

\paragraph{Unsupervised Semantic Segmentation.}  Unsupervised semantic segmentation is less studied in the field, and most of the existing work focuses on non-parametric methods~\cite{Shi2000,Tighe2010}.  Recently, several deep learning-based approaches have been proposed for unsupervised semantic segmentation.  Ji et al.~\cite{Ji2019} proposed a clustering-based approach for image classification and segmentation.  A few self-supervised representation learning approaches~\cite{Hwang2019,Zhang2020a,VanGansbeke2021} were proposed recently, which learn visual representations at the pixel level and use the learned pixel embeddings to segment images.  Our method also belongs to self-supervised representation learning, and the difference between our method and the existing approaches will be discussed in the following section.

\paragraph{Self-supervised Learning.}  Self-supervised learning aims to learn visual representations by predicting one aspect of the input from another, for which various pretext tasks have been used such as predicting image rotations~\cite{Gidaris2018}, solving jigsaw puzzles~\cite{Noroozi2016}, inpainting~\cite{Pathak2016}, etc.  Recently, contrastive learning-based methods~\cite{Chen2020,He2020} have shown great success in learning visual representations in a self-supervised manner.  These methods often use a contrastive loss to map different augmented views of the same input to a common embedding location and distinct from other inputs.  However, these methods mainly learn image-level representations and require curated single-object images.

A few recent approaches learn pixel-level embeddings for the segmentation of complex scenes.  Hwang et al.~\cite{Hwang2019} proposed SegSort, which uses pseudo image segments generated by a contour detector to guide the contrastive learning of pixel embeddings.  Zhang and Maire~\cite{Zhang2020a} group the pseudo image segments hierarchically and use the hierarchical structure to guide the sampling of positive and negative pixel pairs.  Van Gansbeke et al.~\cite{VanGansbeke2021} use saliency masks to learn pixel embeddings.  However, these methods focus on pixels' local affinity information without considering the global semantics in the whole dataset and thus tend to separate objects of the same semantic class in the embedding space.  In this work, we learn visual concepts at both local and global levels only using pseudo image segments.  We leverage different types of relationships between pixels and concepts to improve the pixel embeddings for semantic segmentation.

\paragraph{Visual Concepts.}  Concepts are human interpretable abstractions extracted from images, which are typically represented as image segments with semantic meanings~\cite{Bau2017}.  Visual concepts have been widely used in eXplainable AI to interpret and explain what has been learned by a CNN model.  Most of the existing work along this line of research uses human-specified concepts~\cite{Bau2017,Kim2018} for model interpretation.  Recently, several data-driven approaches have been proposed to extract concepts from images using superpixels~\cite{Ghorbani2019}, prototypes~\cite{Chen2019}, and a dictionary of object parts~\cite{Huang2020}.  Our work bridges the gap between concept extraction and semantic segmentation with pixel embeddings.  On one hand, we propose a new approach for visual concept extraction.  On the other hand, we leverage the visual concepts to improve self-supervised representation learning for semantic segmentation.

\section{Method}
\label{sec:method}

\begin{figure}[t]
\centering
\includegraphics[width=\columnwidth]{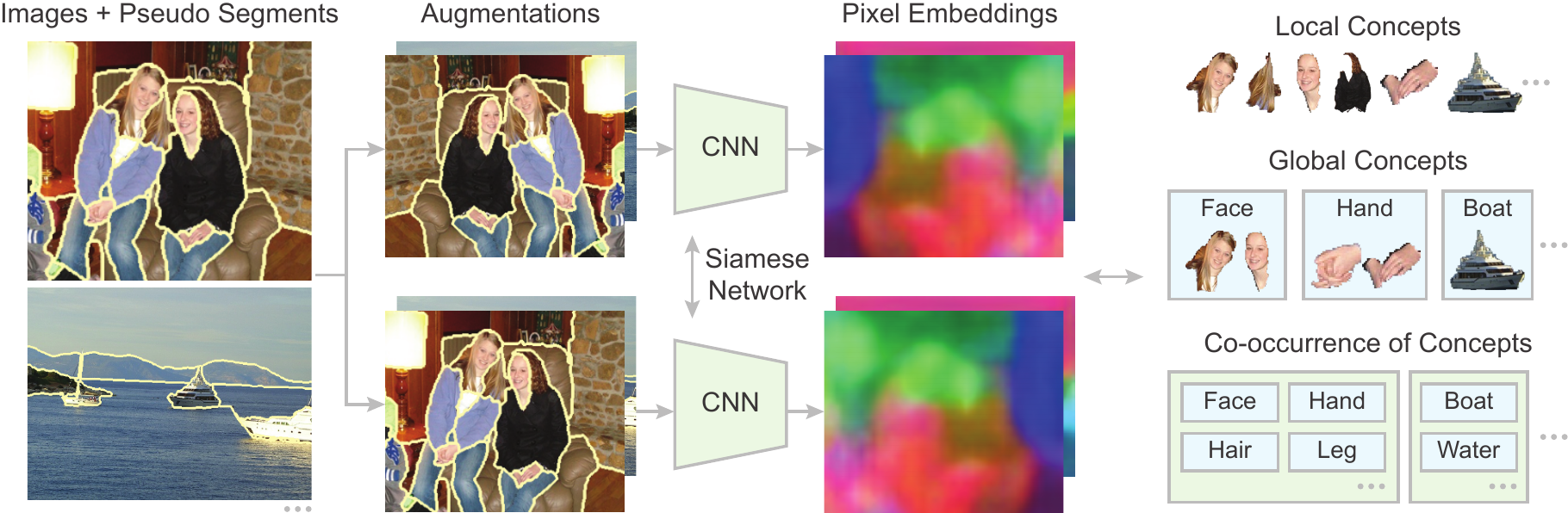}
\caption{Overview of the proposed method.  We learn pixel embeddings for semantic segmentation in a self-supervised setting with pseudo image segments and data augmentations.  We define three types of relationships between pixels and visual concepts to guide the self-supervised learning.}
\label{fig:overview}
\end{figure}

\begin{figure*}[t]
\centering
\includegraphics[width=0.95\textwidth]{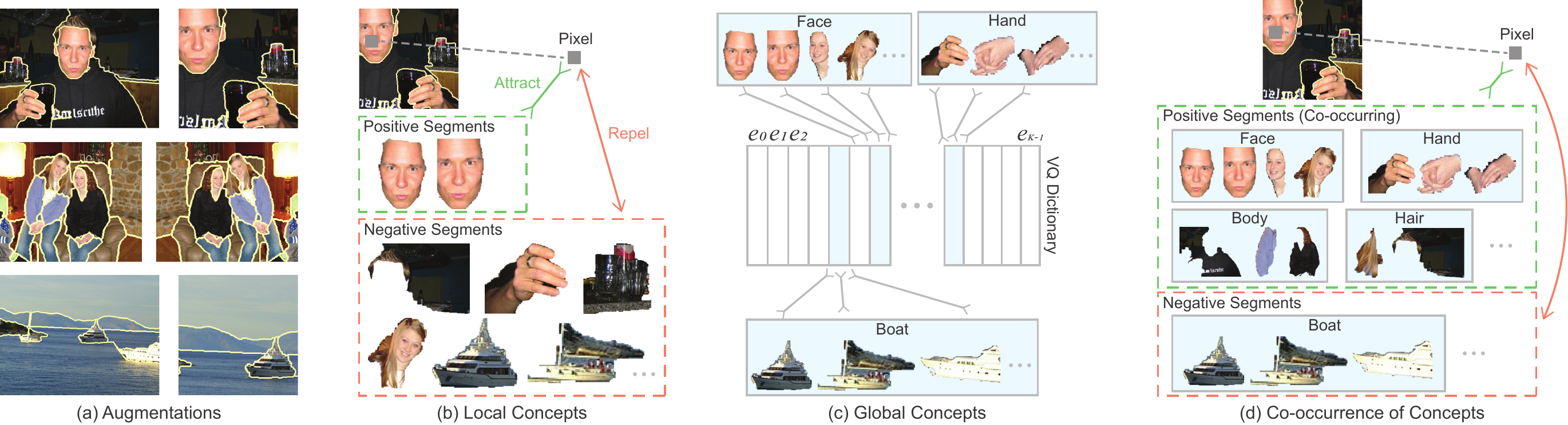}
\caption{Given the augmented views of a set of images with pseudo segments (a), we train pixel embeddings to capture three types of relationships between pixels and visual concepts.  (b) To capture the relationship between pixels and local concepts (i.e., visually coherent regions in each image), we attract the representation of a pixel with the pseudo segments that it belongs to in different augmented views and repel other segments.  (c) We group local concepts with similar feature representations into global concepts with VQ.  The representations of the global concepts form a VQ dictionary that captures the semantic meanings of segment clusters over the entire dataset.  (d) We attract the representations of pixels and segments whose global concepts often co-occur in the same image, such as the human face, body, and hand.}
\label{fig:losses}
\end{figure*}

In this work, we learn a pixel embedding function, namely a convolutional neural network (CNN), for semantic segmentation with a contrastive self-supervised learning framework.  For each pixel $p$, the embedding function $\phi$ generates a feature representation $\mathbf{z}_p$ in an embedding space of dimension $D$, which are then used to derive the semantic segmentation of input images.  Like existing contrastive self-supervised methods~\cite{Chen2020,He2020}, we use data augmentation to improve the learning of visual representations.  Specifically, we generate two augmented views for each image with multiple data augmentation operations such as random cropping and random color jittering.  We then force consistent pixel embeddings between the augmented views.

As pixel-wise labels are not available, our method is based on pseudo image segments of visually coherent regions, which could be derived from super-pixels or contours.  Then we train the pixel embedding function by leveraging important properties of visual concepts learned from the pseudo image segments (Figure~\ref{fig:overview}).  Specifically, we use three types of relationships between pixels and concepts to guide the self-supervised learning, including the relationships between pixels and local concepts, local and global concepts, as well as the co-occurrence of different concepts as detailed below.

\paragraph{Local Concepts.}  We first train the pixel embeddings to conform with visually coherent regions, namely local concepts, in each image.  The idea is that pixels within visually coherent regions should have similar representations in the embedding space~\cite{Hwang2019,Zhang2020a,Ke2021}.  To this end, we define the \textbf{local segmentation loss} $\mathcal{L}_s$ to learn pixel representations with a contrastive representation learning approach following~\cite{Hwang2019}.  Given the augmented views for a batch of images (Figure~\ref{fig:losses}a), we first derive each pixel's positive and negative segments, denoted by $\mathcal{S^+}$ and $\mathcal{S^-}$ respectively, based on the pseudo image segments.  For a pixel $p$, its positive segments include the segments it belongs to in both augmented views, and other segments are negative ones (Figure~\ref{fig:losses}b).  Then, the local segmentation loss is defined as the pixel-to-segment contrastive loss:
\begin{equation}
\mathcal{L}_s(p)=-log\frac{\sum_{s\in{\mathcal{S^+}}}{exp(sim(\mathbf{z}_p, \mathbf{z}_s)\kappa)}}{\sum_{s\in{\mathcal{S^+}\cup\mathcal{S^-}}}{exp(sim(\mathbf{z}_p, \mathbf{z}_s)\kappa)}},
\end{equation}
where $\kappa$ is the concentration constant and $sim(\mathbf{z}_p, \mathbf{z}_s)$ is the cosine similarity between the feature representation $\mathbf{z}_p$ of a pixel $p$ and the feature representation $\mathbf{z}_s$ of a segment $s$.  The feature representation $\mathbf{z}_s$ is defined as the average representation of the pixels within $s$, namely $\mathbf{z}_s=\sum_{p\in{s}}\mathbf{z}_p/|s|$.

\paragraph{Global Concepts.}  We group the local concepts extracted from individual images into global concepts (i.e., clusters of image segments with similar semantic meanings) for the entire dataset.  We introduce the global concepts based on the following observations.  Image segments with similar visual appearance may locate at different regions of the same image or even different images, such as the human faces in Figure~\ref{fig:losses}a.  Since they belong to different local concepts, those segments are considered as negative examples with each other.  Hence, their representations will be pushed away from each other if only considering local concepts, which will eventually hurt the performance of semantic segmentation.  Moreover, as focusing on individual images, local concepts do not capture visual concepts across the entire dataset.

We use VQ~\cite{vandenOord2017} to learn global visual concepts that extract clusters of image segments with similar semantic meanings from the dataset (Figure~\ref{fig:losses}c).  For the segments that belong to the same global concept, the \textbf{VQ loss $\mathcal{L}_v$} makes their representations close to each other.  Specifically, we train a VQ dictionary that contains a set of discrete representations for the concept centers, denoted as $\mathbf{e}_0, \mathbf{e}_1, \dots, \mathbf{e}_{K-1}$, where $K$ is the number of concepts and $\mathbf{e}_i$ is the representation of $i$-th concept.  For each training iteration, we first assign each segment $s$ to the global concept $k$ with the nearest representation:
\begin{equation}
k = argmax_i{sim(\mathbf{z}_s, \mathbf{e}_i)},
\end{equation}
where $sim(\mathbf{z}_s, \mathbf{e}_i)$ is the cosine similarity between representations.  We use the cosine similarity instead of the Euclidean distance used in~\cite{vandenOord2017} is because we learn representations on a hypersphere.  Then, we maximize the cosine similarity between the representations of the segments and the corresponding concepts:
\begin{equation}
\mathcal{L}_v=(1-sim(sg(\mathbf{z}_s), \mathbf{e}_k)) + \beta(1-sim(\mathbf{z}_s, sg(\mathbf{e}_k))),
\end{equation}
where the first half of this function is used to update the VQ dictionary (i.e., cluster centers) as the stop gradient operator $sg$ is applied on the representation of the segments.  Similarly, the second half is used to update the segments' representation while fixing the VQ dictionary.  In addition, we use the commitment constant $\beta$ to control the learning rate of the segments' representation.

The VQ dictionary captures discrete representations with semantic meanings such as human faces and boats for the entire dataset.  It can be used to learn the relationships between different global concepts such as boats and water, which can be exploited to further improve the pixel representations as detailed in the following section.

\paragraph{Co-occurrence of Concepts.}  We utilize the co-occurrence of different global concepts to further improve the pixel embeddings.  The motivation is that global concepts with relevant semantic meanings tend to co-occur in the same image, such as the human face, hair, and body in Figure~\ref{fig:losses}a. However, the representation of relevant concepts (e.g., different human body parts) will be pushed away as they belong to different global concepts if no co-occurrence constraints.

Inspired by~\cite{Ke2021}, we introduce the \textbf{co-occurrence loss $\mathcal{L}_c$} to attract the representations of pixels and segments whose global concepts often co-occur in the same image.  Different from~\cite{Ke2021}, which uses image tags to obtain the co-occurrence, we exploit the VQ dictionary without additional supervision.  Given a pixel $p$, we redefine its positive and negative segments based on the co-occurrence of the global concepts derived from the VQ dictionary.  Specifically, we first determine which global concept the pixel belongs to by looking up the segment containing that pixel in the VQ dictionary.  Then, the pixels positive segments $\mathcal{C}^+$ are defined as the segments that co-occur with the pixel's concept in the same image, and other segments are defined as negative ones $\mathcal{C}^-$.  For example, in Figure~\ref{fig:losses}d, because the pixel belongs to the concept of human face, all image segments that co-occur with human faces are its positive segments, such as different body parts of a person.  The pixel's embedding is then trained based on the contrastive loss similar to the local segmentation loss:
\begin{equation}
\mathcal{L}_c(p)=-log\frac{\sum_{s\in{\mathcal{C^+}}}{exp(sim(\mathbf{z}_p, \mathbf{z}_s)\kappa)}}{\sum_{s\in{\mathcal{C^+}\cup\mathcal{C^-}}}{exp(sim(\mathbf{z}_p, \mathbf{z}_s)\kappa)}},
\end{equation}
where the only difference is the definition of the positive and negative segments.

In the end, we use the three types of relationship between pixels and visual concepts to regularize the self-supervised learning process.  The total loss to train the representation of a pixel $p$ is the weighted combination of the aforementioned three loss terms:
\begin{equation}
\mathcal{L}(p) = \lambda_s\mathcal{L}_s(p) + \lambda_v\mathcal{L}_v(p) + \lambda_c\mathcal{L}_c(p),
\end{equation}
where $\lambda_s$, $\lambda_v$, and $\lambda_c$ are the weights for each loss term.

\section{Experiments}
\label{sec:experiments}

\subsection{Experimental Setup}
\label{sec:experimental_setup}

\paragraph{Datasets.}  We mainly experiment on the Pascal VOC 2012 dataset, which contains 20 object classes and one background class.  Following the prior work~\cite{Hwang2019}, we train networks on the \textit{train\_aug} set with 10,582 images and test on the \textit{val} set with 1,449 images.  For self-supervised pre-training, we use pseudo segmentations generated by HED-owt-ucm~\cite{Hwang2019}, unless otherwise stated.

We also perform experiments on COCO 2017 and DAVIS 2017 to evaluate the generalizability of the learned pixel embeddings.  Note that the pixel embeddings are trained only on the \textit{train\_aug} set of Pascal VOC 2012 without including any images from COCO 2017 or DAVIS 2017.

\paragraph{Training.}  For all the experiments, we use PSPNet~\cite{Zhao2017} with a dilated ResNet-50 backbone as the network architecture.  The backbone is pre-trained on the ImageNet dataset.  For self-supervised pre-training, the hyper-parameters are set as follows.  The embedding dimension is set to 32, and the concentration constant $\kappa$ is set to 10.  For VQ, we use a dictionary of size 512 and set the commitment constant $\beta$ to 0.5.  The weights $\lambda_s$, $\lambda_v$, and $\lambda_o$ of each loss term are set to 1, 2, and 1, respectively.  We train the network on the \textit{train\_aug} set of Pascal VOC 2012 for 5k iterations with a batch size of 8.  We set the initial learning rate to 0.001 and decay it with a poly learning rate policy.  We use additional memory banks to cache the embeddings of the previous 2 batches.  We use the same set of data augmentations as SimCLR~\cite{Chen2020} during training, including random resizing, cropping, flipping, color jittering, and Gaussian blurring.  Note that in the experiments, the same pseudo segment within different augmented views are merged and considered as one segment for loss computation to save time.

\subsection{Results and Analysis}
\label{sec:results}

\paragraph{Pascal VOC 2012 and COCO 2017: Benchmarking Results.}  We evaluate the learned pixel embeddings with two approaches, including $k$-means clustering~\cite{Hwang2019} and linear classification~\cite{Zhang2020a}.  For $k$-means clustering, we follow the procedure of SegSort~\cite{Hwang2019}.  We first segment each image by clustering the pixels based on the embeddings.  We then assign each segment a semantic class label by the majority vote of its nearest neighbors from the training set.  The hyper-parameters are defined as follows.  Each image is clustered into 25 segments through 50 iterations, and 15 nearest neighbors are used for predicting class labels during inference.  For linear classification, we train an additional softmax classifier, namely a 1$\times$1 convolutional layer, while fixing the learned pixel embeddings.  We train the classifier for 60k iterations with a batch size of 16.  The learning rate starts at 0.1 and decays by 0.1 at 20k and 50k iterations.

We compare our method with the three methods that learn pixel embeddings for semantic segmentation include SegSort~\cite{Hwang2019}, MaskContrast~\cite{VanGansbeke2021}, and Hierarchical Grouping~\cite{Zhang2020a}.  Both SegSort and MaskContrast are trained on a backbone network pre-trained on the ImageNet dataset, which is the same as our approach.  For comparison, we train a SegSort model using the same hyper-parameter setting as our approach.  For MaskContrast, we take the best model provided by the authors and evaluate the performance of the model the same as our approach.  Hierarchical Grouping is a slightly different approach, whose goal is to learn pixel embeddings from scratch for general purposes.  For semantic segmentation, Hierarchical Grouping requires training a complete atrous spatial pyramid pooling (ASPP)~\cite{Chen2018b} module based on the learned embeddings and the annotated data.  In contrast, our method as well as SegSort and MaskContrast only train a linear classifier on the annotated data.  For the sake of completeness, we take the resulting numbers from~\cite{Zhang2020a} and compare with other methods.  Note that comparing with methods that produce image-level representations~\cite{Chen2020,He2020} is out the scope of this work.  The interested reader may consult the previous work~\cite{Zhang2020a,VanGansbeke2021} for the comparison results.

\begin{table}[t]
\centering
{\small
\begin{tabular}{c|c|c|c}
  \multirow{2}{*}{Method} & \multirow{2}{*}{Pseudo Seg.} & mIoU & mIoU \\
  & & ($k$-means) & (LC) \\
  \hline
  SegSort & HED & 49.17 & 58.85 \\
  Hierarch. Group. & SE & -- & 46.51 \\
  Hierarch. Group. & HED & -- & 48.82 \\
  MaskContrast & Unsup. Sal. & 49.67 & 62.22$\dagger$ \\
  MaskContrast & Sup. Sal. & 52.35 & 65.08$\dagger$ \\
  \hline
  Ours \textit{w/o $\mathcal{L}_v$\&$\mathcal{L}_c$} & HED & 61.02 & 62.31 \\
  Ours \textit{w/o $\mathcal{L}_c$} & HED & 61.77 & 62.72 \\
  Ours & Felz-Hutt & 60.38 & 61.90 \\
  Ours & HED & \textbf{62.56} & 63.47
\end{tabular}
}
\caption{\textbf{Benchmarking results on Pascal VOC 2012}.  Our method outperforms previous work by a large margin when using $k$-means.  Meanwhile, our approach performs the second best for linear classification.  The method that outperforms our approach is MaskContrast with a supervised saliency estimator trained on annotated data.  $\dagger$ The number is slightly higher than the original paper because we evaluate the models on images at multiple scales.}
\label{tab:pascal}
\end{table}

\begin{table}[t]
\centering
{\small
\begin{tabular}{c|c|c|c}
  \multirow{2}{*}{Method} & \multirow{2}{*}{Pseudo Seg.} & mIoU & mIoU \\
  & & ($k$-means) & (LC) \\
  \hline
  SegSort & HED & 31.02 & 36.07 \\
  MaskContrast & Unsup. Sal. & 27.47 & 39.77 \\
  MaskContrast & Sup. Sal. & 29.82 & 43.19 \\
  \hline
  Ours \textit{w/o $\mathcal{L}_v$\&$\mathcal{L}_c$} & HED & 36.45 & 38.92 \\
  Ours \textit{w/o $\mathcal{L}_c$} & HED & 37.16 & 40.22 \\
  Ours & HED & 38.00 & 40.84
\end{tabular}
}
\caption{\textbf{Benchmarking results on COCO 2017}.  Our method outperforms previous work on $k$-means and performs the second best on linear classification.  The method that outperforms our approach is MaskContrast that trained on annotated data.}
\label{tab:coco}
\end{table}

\begin{figure}[t]
\centering
\includegraphics[width=.9\columnwidth]{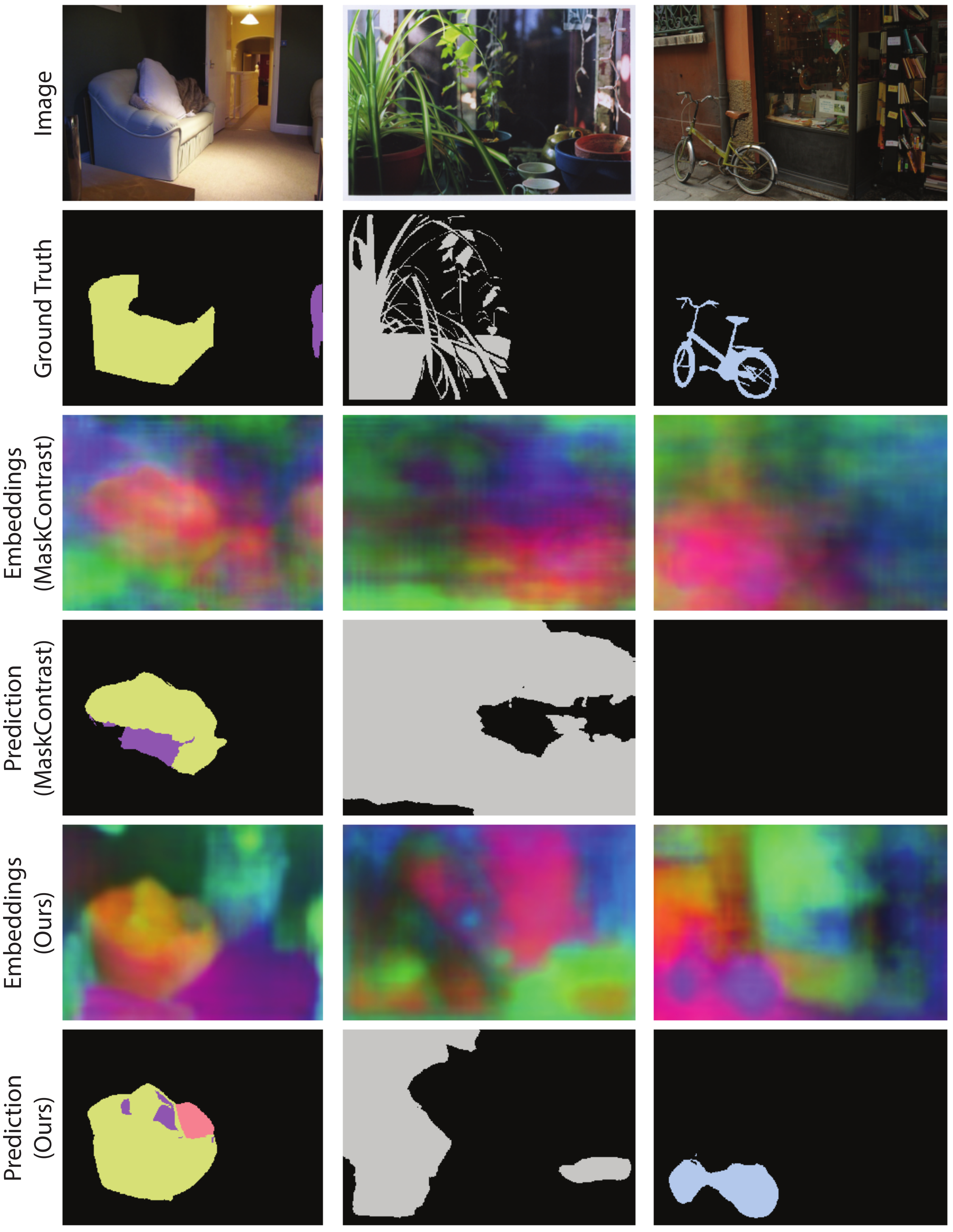}
\caption{\textbf{Visual comparison on PASCAL VOC 2012 validation set.}  Comparing with MaskContrast, our method can produce less noisy pixel embeddings with clear boundaries between different objects, hence generates better semantic segmentation using $k$-means clustering.  Note that the pixel embeddings are visualized by projecting to a 3-dimensional space using PCA.}
\label{fig:pascal_compare}
\end{figure}

\revision{Table~\ref{tab:pascal} shows the benchmarking results of the aforementioned methods on Pascal VOC 2012.  We find that our method outperforms previous methods by a large margin on $k$-means.  The performance of our method on linear classification outperforms MaskContrast with an unsupervised saliency estimator (63.47 vs. 62.22).  Meanwhile, MaskContrast with a supervised saliency estimator has a better performance comparing with ours (65.08 vs. 63.47).  However, it requires training a saliency estimator with a large number of annotated images~\cite{Qin2019}.  Figure~\ref{fig:pascal_compare} compares the learned pixel embeddings and the semantic segmentation generated by $k$-means for our approach and MaskContrast.  We find that our method can produce pixel embeddings with clear boundaries between different objects.  While MaskContrast focuses only on the salient objects and the pixel embeddings are less informative in other regions.  We also examine the effects of the three losses by introducing them one by one (shown in Table 1).  With global concepts, the model is improved by 0.75\% on $k$-means.  With concepts co-occurrence, the model is further improved by 0.8\%.  Similar results can also be found in linear classification.  In addition, our method can achieve comparable performance to previous work even with pseudo segments generated by a non-parametric contour detector~\cite{Felzenszwalb2004}.  \textbf{Detailed ablation study results are included in the supplementary.}}

\revision{Table~\ref{tab:coco} shows the benchmarking results on COCO 2017.  Again, our method outperforms previous work on $k$-means and performs the second best on linear classification.}

\paragraph{Pascal VOC 2012: Analysis of Visual Concepts.}  We further analyze the learned visual concepts, whose quality plays an important role in the performance of our approach.  To this end, we segment each image based on the discrete representation of the global concepts, namely the VQ dictionary.  Given an input image, we first generate the embedding of each pixel.  Then we cluster the pixels into segments based on the pixel embeddings.  In the end, we map each segment into one of the concepts based on their distance to the concepts.  We also merge the segments with the same concept if they are connected.  Based on the extracted image segments, the visual concepts are evaluated from different aspects.

\begin{figure}[t]
\centering
\includegraphics[width=0.95\columnwidth]{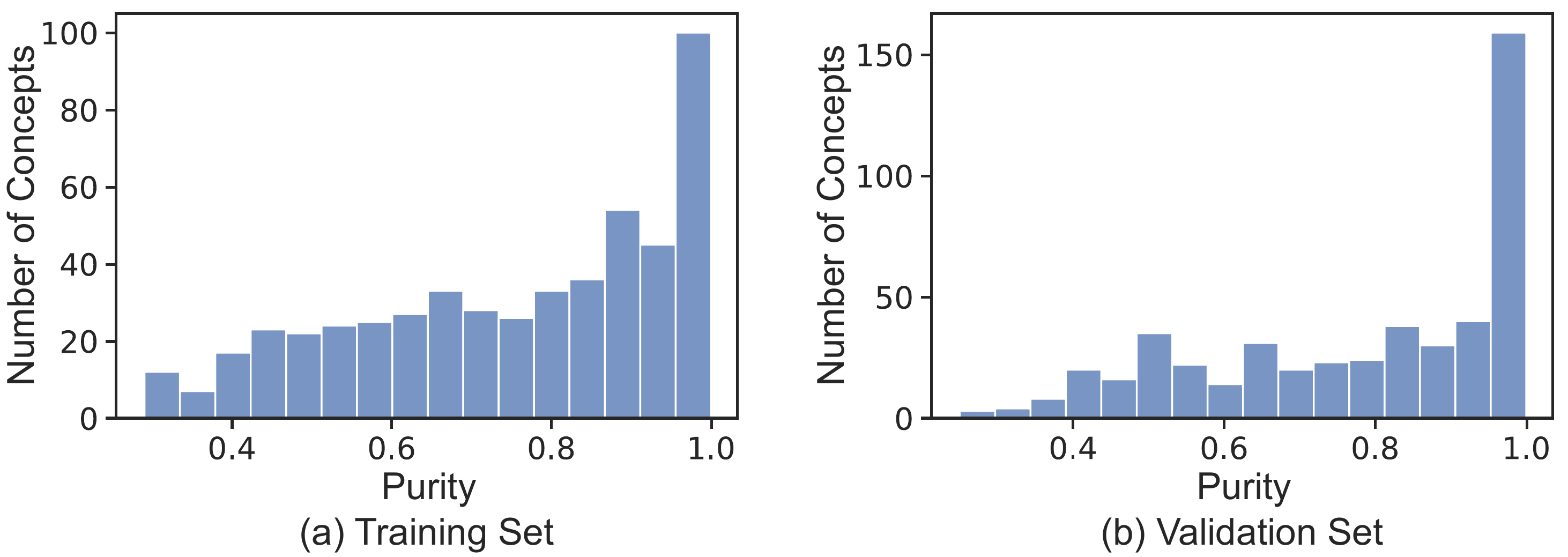}
\caption{\textbf{Distribution of the purity for the visual concepts on the (a) training and (b) validation set.}  More than half of the concepts have the purity higher than 80\% on both sets.}
\label{fig:purity_hist}
\end{figure}

To evaluate the visual concepts, we first answer the question that if each visual concept captures image features with the same semantic meaning.  To this end, we define the \textit{purity} of each concept as follows.  For each concept, we collect the corresponding image segments and assign each image segment with a class label based on the ground truth.  Then we calculate the percentage of the image segments that belong to the majority class of the current concept.  The distribution of the purity for the visual concepts is shown in Figure~\ref{fig:purity_hist}.  We find that more than half of the concepts have the purity higher than 80\% on both the training and validation set.

\begin{figure}[t]
\centering
\includegraphics[width=0.95\columnwidth]{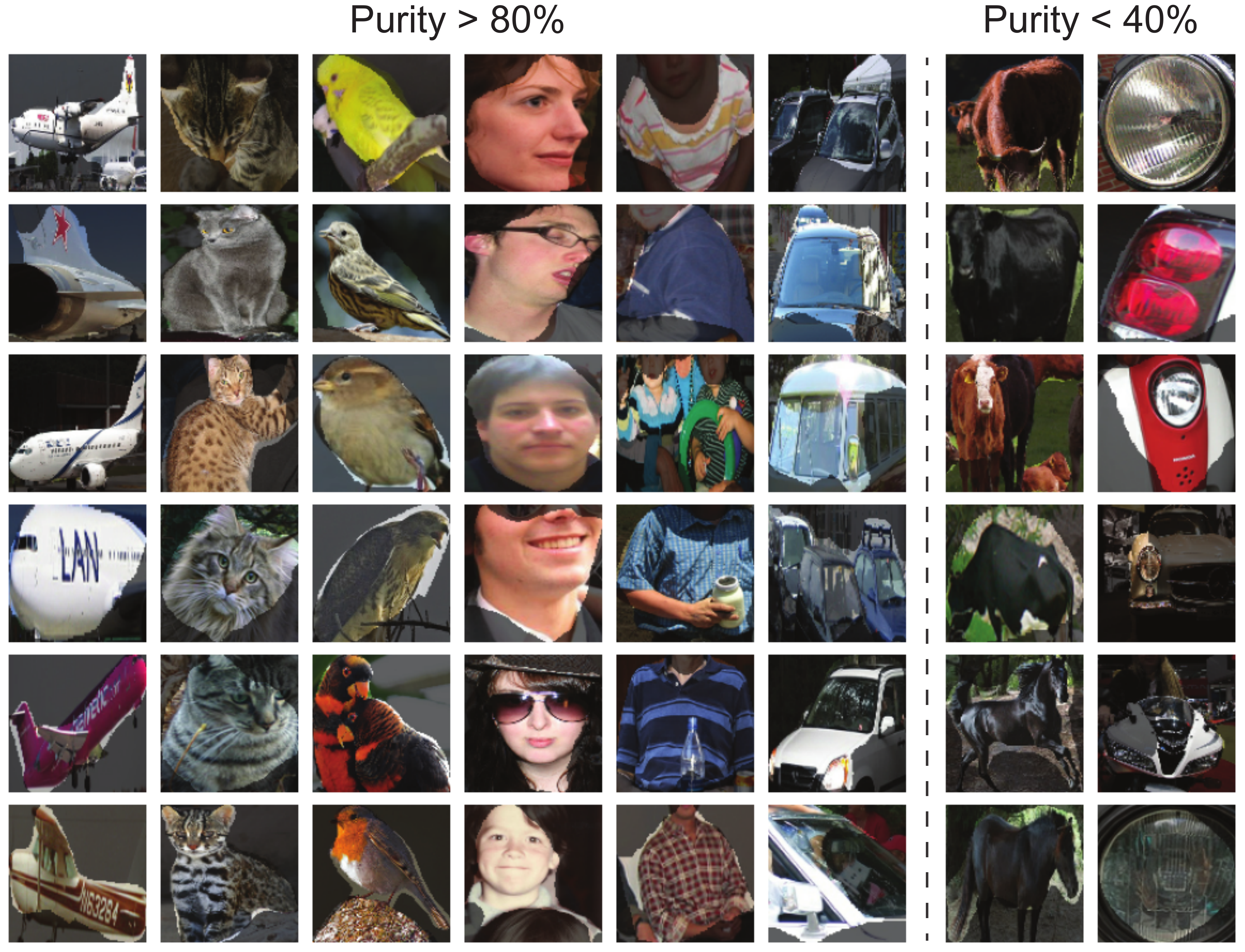}
\caption{\textbf{Image segments randomly sampled from a few concepts (one concept per column)}.  Concepts of high purity capture image segments with clear semantic meanings.  Moreover, fine-grained visual concepts can be extracted from the dataset using our method, such as human face, cloth with stripes, and car window.  Concepts of low purity can also capture image features with similar meanings such as farm animals and round structures in vehicles.}
\label{fig:vq_samples}
\end{figure}

Figure~\ref{fig:vq_samples} visualizes image segments randomly sampled from a few concepts (one concept per column).  The first 6 columns are from concepts with purity higher than 80\%, and the last two columns are from concepts with purity lower than 40\%.  We find that the visual concepts of high purity contain image segments with clear semantic meanings such as airplane, cat, and bird.  Moreover, our method can extract fine-grained visual concepts such as human face, cloth with stripes, and car window.  For the concepts with low purity, the image segments still capture similar features such as farm animals and round structures in vehicles.  However, without annotations, the network has difficulty differentiating horses and cows or determining the type of vehicles.

\paragraph{DAVIS 2017: Instance Mask Tracking.}  We evaluate the generalizability of our method on instance mask tracking over the DAVIS 2017 validation set, where the instance masks at the first frame are given for each video.  We propagate the instance masks to the rest of the frames based on the similarity between pixel embeddings following the method proposed in~\cite{Zhao2017}.  The performance is measured by the region similarity $\mathcal{J}$ (IoU) and the contour-based accuracy $\mathcal{F}$.

\begin{table}[t]
\centering
\begin{tabular}{c|c|c}
  Method & $\mathcal{J}$(Mean)$\uparrow$ & $\mathcal{F}$(Mean)$\uparrow$ \\
  \hline
  MaskTrack (fine-tuned)$\dagger$ & 51.2 & 57.3 \\
  OSVOS (fine-tuned)$\dagger$ & 55.1 & 62.1 \\
  \hline
  MaskTrack-B & 35.3 & 36.4 \\
  OSVOS-B & 18.5 & 30.0 \\
  \hline
  Video Colorization & 34.6 & 32.7 \\
  CycleTime & 41.9 & 39.4 \\
  mgPFF & 42.2 & 46.9 \\
  Hierarch. Group. & 47.1 & 48.9 \\
  MaskContrast (Sup.) & 34.3 & 36.7 \\
  \hline
  Ours (Felz-Hutt) & 46.3 & 49.0 \\
  Ours & \textbf{50.4} & \textbf{53.9} \\
\end{tabular}
\caption{\textbf{Performance of instance mask tracking on DAVIS 2017 validation set.}  The performance is measured by the region similarity $\mathcal{J}$ and the contour-based accuracy $\mathcal{F}$.  Our method outperforms recent supervised and unsupervised methods on both metrics.  $\dagger$ indicates the models are fine-tuned on the first frame of the test videos.}
\label{tab:davis}
\end{table}

We compare our method with recent supervised and unsupervised methods in Table~\ref{tab:davis}.  The supervised methods MaskTrack-B~\cite{Perazzi2017} and OSVOS-B~\cite{Caelles2017} train models with ImageNet pre-training and annotated masks.  MaskTrack~\cite{Perazzi2017} and OSVOS~\cite{Caelles2017} further fine-tune the models on the first frame of the test video.  Comparing to the supervised methods, our method outperforms MaskTrack-B and OSVOS-B by a large margin without using any annotated masks.  Also, our method achieves more than 91\% and 87\% performance of the fine-tuned models in terms of the region similarity $\mathcal{J}$ and contour accuracy $\mathcal{F}$, respectively.  Our method also outperforms recent video-based~\cite{Vondrick2018,Wang2019a,Kong2019} and image-based~\cite{Zhang2020a,VanGansbeke2021} unsupervised approaches by more than 3\% and 5\% in $\mathcal{J}$ and $\mathcal{F}$, respectively.  Moreover, our method can achieve comparable performance to previous work even with pseudo segments generated by a non-parametric contour detector~\cite{Felzenszwalb2004}.  Due to limited space, visual results are included in the supplementary.

\section{Conclusion}
\label{sec:conclusion}

We propose a novel unsupervised semantic segmentation method based on self-supervised representation learning at the pixel level.  Our method uses three types of relationships between pixels and visual concepts to regularize the self-supervised representation learning and hence improve the learned pixel embeddings.  We demonstrate the accuracy and generalizability of the learned pixel embeddings on PASCAL VOC2012, COCO 2017, and DAVIS 2017.

\bibliographystyle{named}
\bibliography{main}

\end{document}


\maketitle

\appendix

\section{Supplementary Material}

In this supplementary material, we include:
\begin{itemize}
  \item ablation study regarding hyper-parameters;
  \item per-class results on Pascal VOC 2012 validation set;
  \item visualization of the global concepts;
  \item visualization of the pixel embeddings and the instance masks on DAVIS-2017 validation set.
\end{itemize}

\subsection{Ablation Study}

\revision{We conduct experiments for the ablation study to understand how the hyper-parameters affect the performance of our method, including the weights of the losses, the number of global concepts, the commitment constant $\beta$, and the memory bank size.}

\revision{\paragraph{Weights of the losses.}  We fix $\lambda_s$ to 1 and vary $\lambda_v$ and $\lambda_c$ to perform the ablation experiments.  The results are shown in Table~\ref{tab:vq_weight}.  We find that our model performs well (i.e., outperforms the model that only uses local concepts by a large margin) when $\lambda_v$ is greater than 1 and $\lambda_c$ is smaller than 2.}

\begin{table}[h]
\centering
\setlength{\tabcolsep}{2.5pt}
{\small
\begin{tabular}{c|c|cc}
  \multirow{2}{*}{$\lambda_v$} & \multirow{2}{*}{$\lambda_c$} & mIoU & mIoU \\
  & & ($k$-means) & (LC) \\
  \hline
  0.5 & 1 & 61.58 & 62.86 \\
  1 & 1 & 62.08 & 63.06 \\
  1.5 & 1 & 62.22 & 63.22 \\
  2 & 1 & 62.56 & 63.47 \\
  2.5 & 1 & 62.62 & 63.83 \\
  3 & 1 & 62.27 & 63.47 \\
\end{tabular}
\quad
\begin{tabular}{c|c|cc}
  \multirow{2}{*}{$\lambda_v$} & \multirow{2}{*}{$\lambda_c$} & mIoU & mIoU \\
  & & ($k$-means) & (LC) \\
  \hline
  2 & 0.5 & 62.10 & 62.88 \\
  2 & 1 & 62.56 & 63.47 \\
  2 & 1.5 & 61.98 & 64.04 \\
  2 & 2 & 61.62 & 63.31 \\
  2 & 2.5 & 61.07 & 63.41 \\
  2 & 3 & 60.97 & 63.21 \\
\end{tabular}
}
\caption{Experiments on different weights of loss terms.}
\label{tab:vq_weight}
\end{table}

\paragraph{Number of global concepts.}  We train our model on Pascal VOC 2012 \textit{train\_aug} set with 32, 64, 128, 256, 512, and 1024 global concepts.  We test the trained models with $k$-means clustering and nearest neighbors retrieval on the validation set of Pascal VOC 2012.  The experiment results are shown in Table~\ref{tab:nconcepts}.  We find that the model does not perform well if the number of global concepts is smaller than 128.  The reason could be that the model aggregates image segments with different semantic meanings into the same global concept as there are only a small number of global concepts.

\begin{table}[h]
\centering
\begin{tabular}{c|c}
  Num of Global Concepts & mIoU \\
  \hline
  32 & 60.90 \\
  64 & 61.45 \\
  128 & 61.85 \\
  256 & 62.07 \\
  512 & 62.56 \\
  1024 & 62.20 \\
\end{tabular}
\caption{Model's performance with respect to the number of global concepts.  To obtain a good performance, there should be enough global concepts (e.g., more than 256 concepts).}
\label{tab:nconcepts}
\end{table}

\paragraph{Commitment constant $\beta$.}  The commitment constant $\beta$ is used to control how much the global concepts affect the pixel embeddings.  Higher $\beta$ means the pixel embeddings are more concentrate around the learned global concepts.  We train our model with different values of $\beta$ on Pascal VOC 2012 \textit{train\_aug} set and test the trained models with $k$-means clustering and nearest neighbors retrieval on the validation set of Pascal VOC 2012.  We can see that the model performs better if the commitment constant $\beta$ is greater than 0.3, which indicates that regularizing the pixel embeddings based on the global concepts can benefit the model's performance.  Note that although the choice of the hyper-parameters could affect the model's performance, our model still outperforms previous work by more than 5\% with the hyper-parameters used in the ablation study.

\begin{table}[h]
\centering
\begin{tabular}{c|c}
  Commitment Constant $\beta$ & mIoU \\
  \hline
  0.1 & 61.64 \\
  0.2 & 61.46 \\
  0.3 & 62.03 \\
  0.4 & 62.34 \\
  0.5 & 62.56 \\
  0.6 & 62.29 \\
  0.7 & 62.10 \\
\end{tabular}
\caption{Model's performance with respect to the commitment constant $\beta$.  The model performs better if the commitment constant $\beta$ is greater than 0.3, which means regularizing the pixel embeddings based on the global concepts can benefit the model's performance.}
\label{tab:commitment}
\end{table}

\begin{table*}[t]
\centering
\resizebox{\textwidth}{!}{
\begin{tabular}{c|c|ccccccccccccccccccccc|c}
  Method & Pseudo Seg. & Background & Aero & Bike & Bird & Boat & Bottle & Bus & Car & Cat & Chair & Cow & Table & Dog & Horse & MBike & Person & Plant & Sheep & Sofa & Train & TV & mIoU \\
  \hline
  SegSort      & HED         & 88.30          & 67.52          & 29.34          & 67.82          & 39.88          & 49.45          & 61.94          & 64.88          & 74.43          &  5.85          & 26.41          & 16.33          & 67.03          & 44.06          & 63.28                & 71.21          & 13.28          & 52.15          & 22.65          & 58.15          & 48.63          & 49.17 \\
  MaskContrast & Unsup. Sal. & 84.27          & 68.32          & 23.26          & 49.78          & 42.44          & 54.99          & \textbf{82.21} & 67.02          & 62.57          &  7.15          & 50.38          & 33.96          & 59.29          & 49.66          & 60.12                & 52.37          & 20.63          & 55.47          & 21.28          & 63.33          & 34.64          & 49.67 \\
  MaskContrast & Sup. Sal.   & 86.00          & 62.99          & 23.63          & 56.08          & 45.28          & 61.31          & 74.68          & 71.33          & 64.89          &  6.70          & 46.55          & 35.21          & 63.80          & 46.67          & 58.83                & 58.03          & 32.53          & 60.07          & 29.69          & 67.17          & 47.95          & 52.35 \\
  \hline
  Ours & HED                 & \textbf{91.02} & \textbf{80.47} & \textbf{33.93} & \textbf{80.42} & \textbf{59.93} & \textbf{67.15} & 81.94          & \textbf{74.00} & \textbf{83.25} & \textbf{19.00} & \textbf{52.45} & \textbf{50.85} & \textbf{76.12} & \textbf{54.41} & \textbf{66.89} & \textbf{75.63} & \textbf{37.97} & \textbf{65.57} & \textbf{34.47} & \textbf{70.07} & \textbf{58.34} & \textbf{62.56}
\end{tabular}
}
\caption{Per-class results on Pascal VOC 2012 validation set using $k$-means clustering and nearest neighbors retrieval.  Our method outperforms previous work consistently.}
\label{tab:pascal_perclass_kmeans}
\end{table*}

\begin{table*}[t]
\centering
\resizebox{\textwidth}{!}{
\begin{tabular}{c|c|ccccccccccccccccccccc|c}
  Method & Pseudo Seg. & Background & Aero & Bike & Bird & Boat & Bottle & Bus & Car & Cat & Chair & Cow & Table & Dog & Horse & MBike & Person & Plant & Sheep & Sofa & Train & TV & mIoU \\
  \hline
  SegSort & HED              & 89.82          & 74.26          & \textbf{35.34} & 74.87          & 54.60          & 63.75          & 71.30          & 68.07          & 78.57          & 18.01          & 53.91          & 30.15          & 70.54          & 54.01          & 68.43          & 73.59          & 41.22          & 62.41          & 25.45          & 65.13          & \textbf{62.40} & 58.85 \\
  MaskContrast & Unsup. Sal. & 90.10          & 85.38          & 34.22          & 76.46          & 63.57          & 67.93          & 84.22          & 75.00          & 78.59          & 14.20          & 66.29          & 40.88          & 73.55          & 62.00          & 71.86          & 70.48          & 30.02          & 71.44          & \textbf{30.63} & 72.22          & 47.60          & 62.22 \\
  MaskContrast & Sup. Sal.   & \textbf{91.26} & \textbf{87.23} & 32.98          & 79.56          & \textbf{65.68} & \textbf{68.13} & \textbf{84.48} & \textbf{77.47} & 81.31          & 20.59          & \textbf{68.22} & \textbf{44.82} & 75.57          & \textbf{65.11} & \textbf{72.28} & 73.90          & 40.87          & \textbf{76.03} & 30.29          & \textbf{76.48} & 54.48          & \textbf{65.08} \\
  \hline
  Ours & HED                 & 90.73          & 78.06          & 35.28          & \textbf{80.78} & 56.09          & 68.04          & 81.80          & 73.73          & \textbf{85.02} & \textbf{24.75} & 62.41          & 42.07          & \textbf{76.92} & 61.04          & 67.15          & \textbf{76.85} & \textbf{41.83} & 67.73          & 29.25          & 71.95          & 61.44          & 63.47
\end{tabular}
}
\caption{Per-class results on Pascal VOC 2012 validation set using linear classification.  Our method outperforms most of the previous work except MaskContrast with a salience estimator trained on a large number of annotated images.}
\label{tab:pascal_perclass_lc}
\end{table*}

\begin{figure*}[t]
\centering
\includegraphics[width=\textwidth]{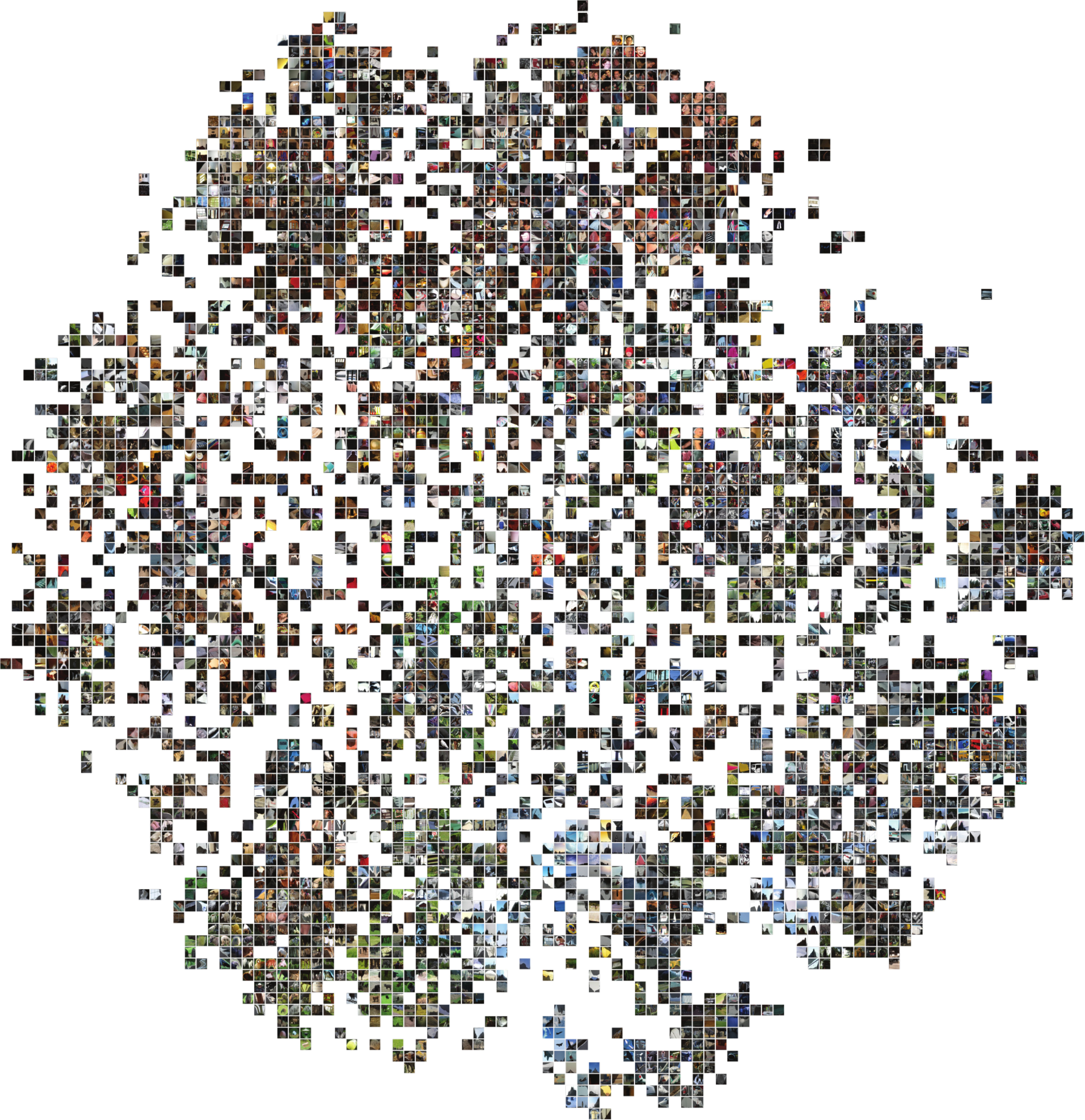}
\caption{t-SNE visualization of image segments sampled from global concepts based on the segments' latent representation.  Best viewed with zoom-in.}
\label{fig:tsne}
\end{figure*}

\begin{figure*}[t]
\centering
\includegraphics[width=\textwidth]{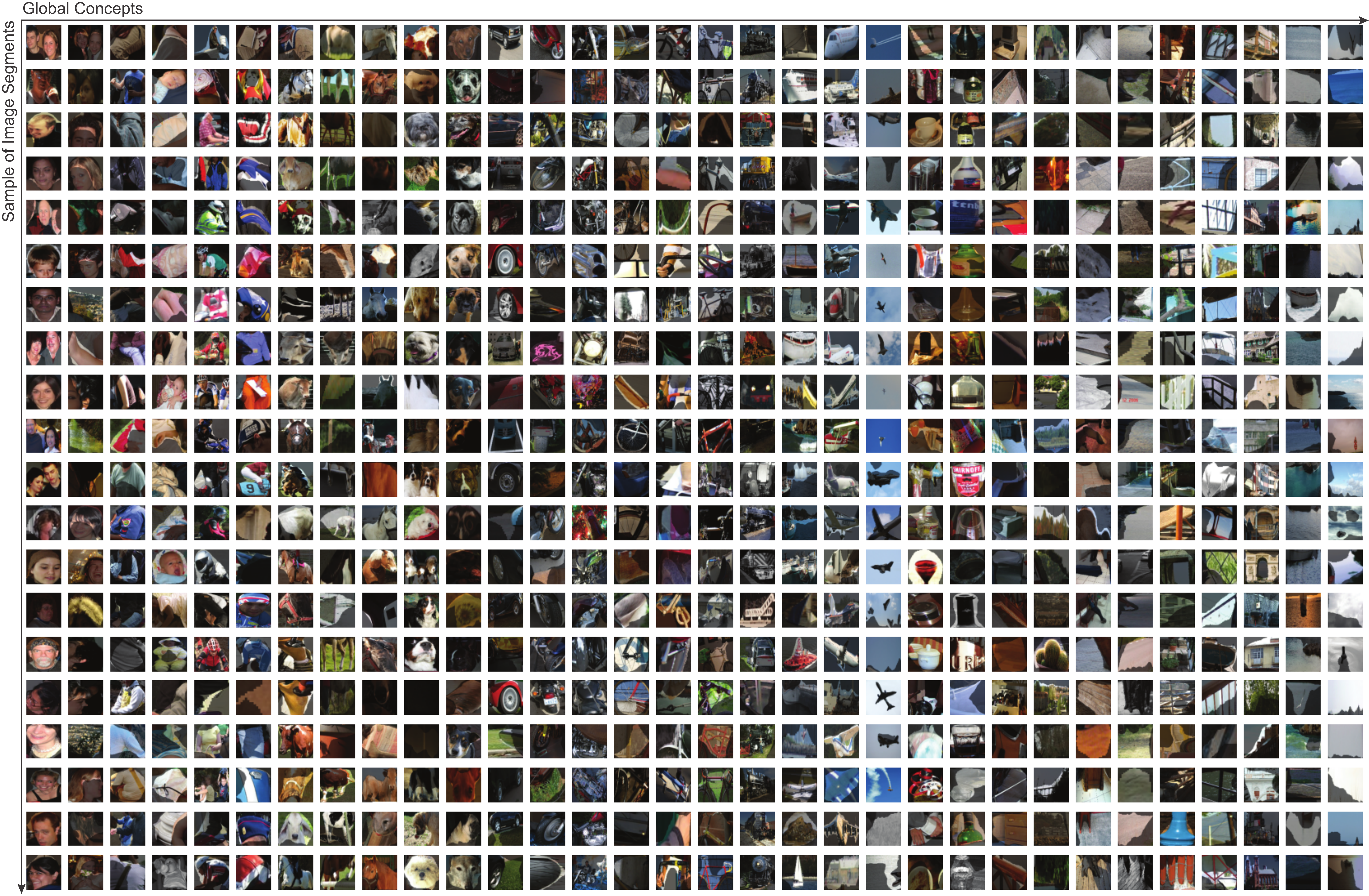}
\caption{Image segments sampled from 32 concepts (one concept per column).  Fine-grained visual concepts representing different object parts or specific scenarios can be extracted from the data using our method.}
\label{fig:dic_vis}
\end{figure*}

\begin{figure}[t]
\centering
\includegraphics[width=0.95\columnwidth]{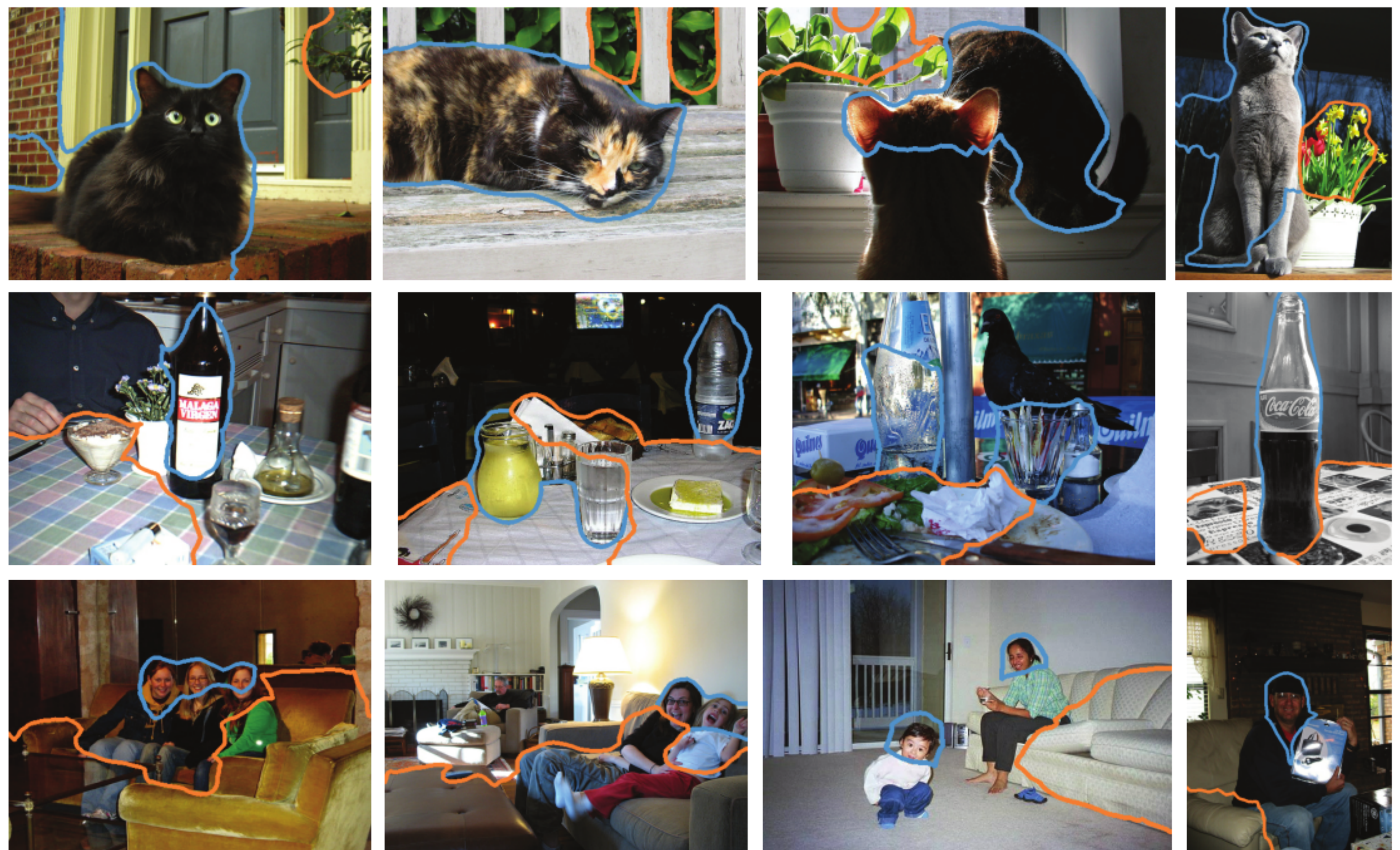}
\caption{Image retrieval based on the visual concepts.  By associating the learned visual concepts, we can create sophisticated scenarios for image retrieval, such as cat next to plants, bottles on a table, or people sitting on a sofa.}
\label{fig:query}
\end{figure}

\begin{figure*}[t]
\centering
\includegraphics[width=\textwidth]{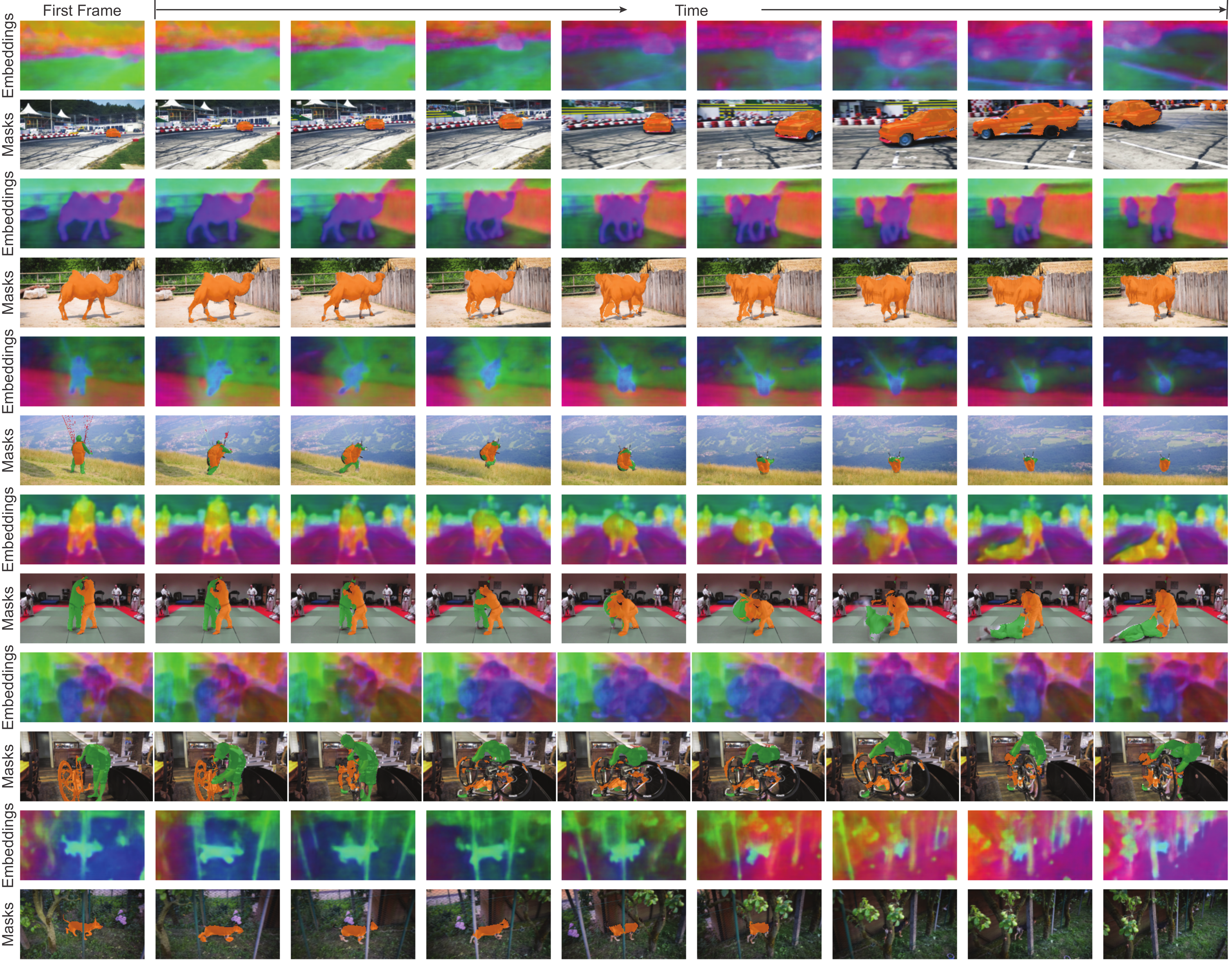}
\caption{Visualization of the pixel embeddings and instance masks produced by our method on DAVIS-2017 validation set.}
\label{fig:davis_more}
\end{figure*}

\revision{\paragraph{Memory bank size.}  By training our model with different memory bank size, we find that our method performs better with a bigger memory bank (e.g., mIoUs are 59.96, 61.62, and 62.56 for memory bank of 0, 1, and 2 batches, respectively).  Similar findings were also discovered in the previous work (Hwang et al. 2019).  In the main paper, we use a memory bank of 2 batches (8 images per batch) for all the baselines except hierarchical grouping (Zhang and Maire 2020), which uses a much bigger batch size (i.e., 70 images per batch).}

\subsection{Per-class Results on Pascal VOC 2012}

Table~\ref{tab:pascal_perclass_kmeans} shows per-class results on Pascal VOC 2012 validation set using $k$-means clustering and nearest neighbors retrieval.  Our method outperforms previous work consistently.  Especially on class \textit{aero}, \textit{bird}, \textit{chair}, and \textit{table}, our method outperforms previous work by more than 10\% on IoU.

Table~\ref{tab:pascal_perclass_lc} shows per-class results on Pascal VOC 2012 validation set using linear classification.  Our method outperforms most of the previous work on most of the classes.  The only exception is MaskContrast with a salience estimator trained on a large number of annotated images, which outperforms our method on approximately 2/3 of the classes.

\subsection{Visualization of the Concepts}

We first use t-SNE to visualize the image segments extracted from Pascal VOC 2012 validation set based on the global concepts.  We project the latent representation of the image segments to a 2-dimensional space and use the 2-dimensional embeddings to layout the image segments.  The visualization is shown in Figure~\ref{fig:tsne}.  We observe that the image segments form clear visual groups in the latent space.  For example, human faces are clustered on the top right.  Buses are placed on the right side.  Farm animals and grasslands are placed on the bottom left.

We then visualize the learned visual concepts by sampling a set of image segments from each concept as shown in Figure~\ref{fig:dic_vis}.  We can see that our method can extract fine-grained visual concepts representing different object parts (e.g., human faces, car tires, and animal feet) and sub-classes (e.g., babies, cups, and fences).  In addition, our method can learn concepts for specific scenarios, such as a human wearing a helmet or an airplane flying in the sky.

By associating the learned visual concepts, we can create sophisticated scenarios for image retrieval and data analysis.  For example, in Figure~\ref{fig:query}, we use the relationships (e.g., relative positions) between different concepts to retrieve images that contain cat next to plants, bottles on a table, or people sitting on a sofa.

\subsection{Visualization of Pixel Embeddings on DAVIS-2017}

Figure~\ref{fig:davis_more} visualizes the pixel embeddings and the instance masks produced by our method on DAVIS-2017 validation set.  For each video, we visualize the results till late frames.  We find that our method can produce relatively consistent embeddings for long sequences of video frames and various types of objects such as cars, humans, and animals.  Meanwhile, our method may lose track of objects that are heavily occluded such as the last row in Figure~\ref{fig:davis_more}.
